\documentclass[sigplan,nonacm]{acmart}
\usepackage{multirow}
\usepackage{graphicx}
\usepackage{subcaption}
\usepackage{enumitem}
\usepackage{algorithm}
\usepackage{algpseudocode}
\AtBeginDocument{%
  \providecommand\BibTeX{{%
    \normalfont B\kern-0.5em{\scshape i\kern-0.25em b}\kern-0.8em\TeX}}}

\PassOptionsToPackage{table}{xcolor}

\definecolor{LightPurple}{RGB}{237,229,255}
\definecolor{DarkPurple}{RGB}{90,40,140}

\newcommand{\hlpurple}[1]{\colorbox{LightPurple}{\strut #1}}

\setcopyright{none}

\begin{document}
\title{Harness–Model Collaboration for Long-Horizon Continual Learning}

\author{%
  Zhen Bi\textsuperscript{1,*},
  Zhizhi Peng\textsuperscript{1,*},
  Yan Wang\textsuperscript{2},
  Xueshu Chen\textsuperscript{1},
  Haofei Zhang\textsuperscript{3},\\
  Yaobo Zhang\textsuperscript{4},
  Zeyu Yang\textsuperscript{1},
  Qing Shen\textsuperscript{1},
  Jungang Lou\textsuperscript{1,\textdagger}
}

\affiliation[obeypunctuation=true]{%
  \institution{%
    \textsuperscript{1}Huzhou Normal University \quad
    \textsuperscript{2}Alibaba Group \quad
    \textsuperscript{3}Zhejiang University \quad
    \textsuperscript{4}Ningxia University\\[4pt]
    {\small
      \textsuperscript{*}These authors contributed equally to this research. \quad
      \textsuperscript{\textdagger}Corresponding author.
    }
  }%
  \country{China}%
}

\renewcommand{\shortauthors}{Bi et al.}

\begin{abstract}
Long-horizon agents continuously encounter reusable procedural knowledge, yet existing systems typically rely either on model adaptation or external memory, trading persistent learning for rapid reuse. This limitation is especially important for Web agents that repeatedly execute related multi-step workflows. We propose \textbf{HaMo}, a Harness--Model collaborative framework that combines fast non-parametric experience accumulation with slow parametric adaptation. HaMo distills trajectories into strategy-level experience cards for immediate retrieval, while an experience-conditioned continual learning objective trains the model to interpret, ground, and execute retrieved strategies across changing contexts. A retrieval-coverage-based selection mechanism further identifies compact and complementary trajectories for parameter updates. Experiments show that HaMo outperforms harness-only and model-only adaptation under within-domain and cross-domain shifts, improving both retention and adaptation. These results highlight the importance of jointly preserving procedural experience and internalizing the competence to transfer and execute it across tasks.
\end{abstract}




\maketitle

\section{Introduction}

An agent that completes a long-horizon task today should become better at solving related tasks tomorrow.
Long-horizon interactions naturally produce reusable procedural knowledge~\cite{zhao2024expel,majumder2023clin}, 
yet agents without explicit cross-task learning mechanisms treat each task as an isolated episode and cannot systematically accumulate such experience over time. 
This limitation is particularly important for Web agents, which repeatedly perform multi-step workflows such as information seeking, product search, and transactional tasks~\cite{deng2023mind2web,he2024webvoyager,drouin2024workarena}, where knowledge acquired from previous interactions could directly benefit future ones.

\begin{figure}[t]
    \centering
    \includegraphics[width=1.0\linewidth]{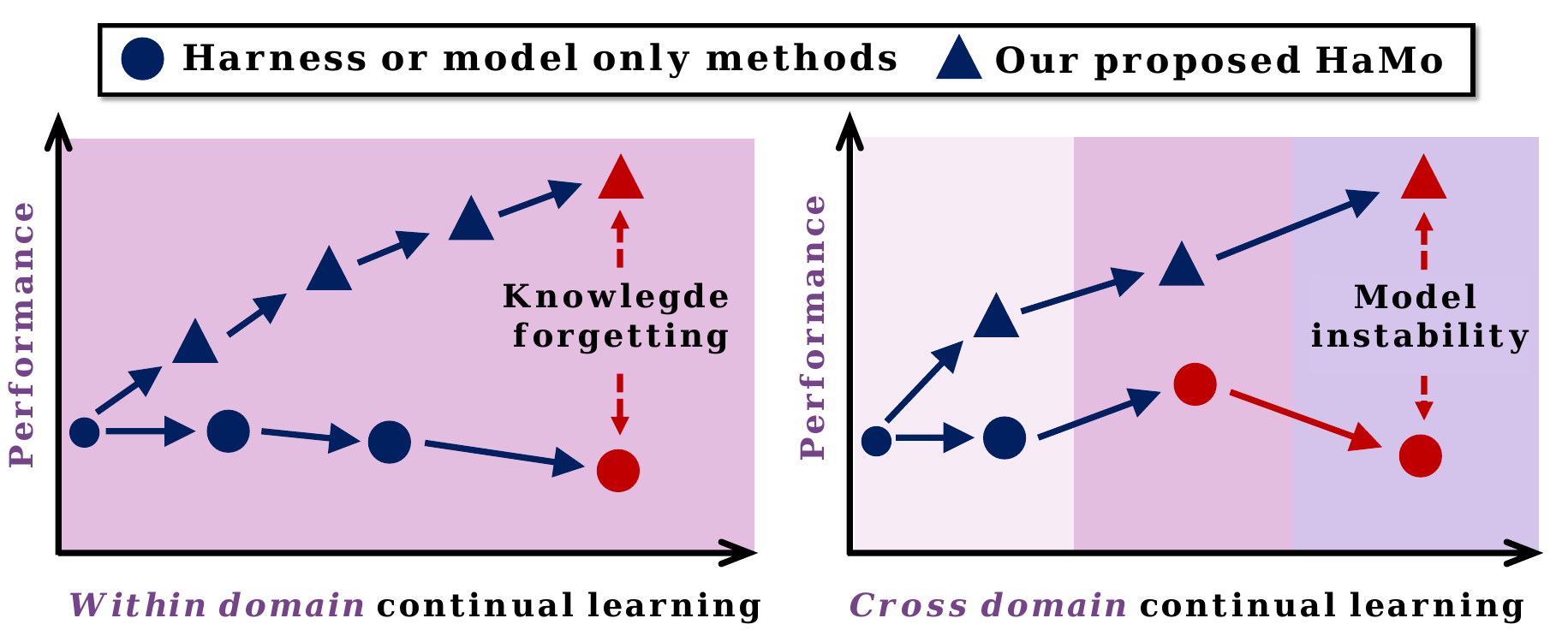}
    \caption{
    Motivation for our method in continual learning for long-horizon agents. 
    Harness- and model-only methods either enable fast experience reuse but tend to plateau, or suffer from instability and forgetting.
    HaMo combines both mechanisms to achieve steady improvement within domains and better retention across domain shifts. 
    }
    \label{fig:intro}
\end{figure}


A natural solution is \emph{model-level adaptation}, where trajectories from previous tasks are used to update model parameters~\cite{chen2023fireact,xi2024agentgym,putta2024agentq}. Such updates can turn past interactions into persistent behavioral competence, but frequent adaptation is costly and may interfere with previously acquired capabilities, especially under sequential distribution shifts~\cite{wang2023continualsurvey,luo2023forgetting}. An alternative is \emph{harness-level adaptation}, where useful experience is stored externally and retrieved for future tasks~\cite{park2023generative,zheng2024synapse}. This enables newly acquired knowledge to affect behavior immediately without retraining, but the model itself does not necessarily become better at interpreting and executing the retrieved experience. Neither mechanism alone fully addresses continual learning in long-horizon agents (shown in Figure \ref{fig:intro}).

We argue that these two forms of adaptation are naturally complementary because they operate at different timescales. 
A strategy discovered from a single trajectory may already be useful for future tasks, while still being too specific to justify an immediate parameter update. 
Such knowledge can first be stored externally and reused when similar situations arise. 
As related strategies recur across tasks, they provide stronger supervision for improving the model itself. For long-horizon agents, this process is not simply about memorizing past solutions:
the key challenge is to internalize \emph{procedural competence}, i.e., the ability to interpret high-level strategies, ground them in changing environments, and translate them into appropriate actions.

Motivated by this observation, \textbf{we propose HaMo, a Harness-Model collaborative framework for continual learning in long-horizon agents}. 
Our method separates adaptation into two tightly coupled processes. At the harness level, each trajectory is distilled into a compact \emph{experience card} containing the task context, initial conditions, and a strategy-level solution outline. These cards are stored in a persistent non-parametric memory, making newly acquired procedural knowledge immediately available to future tasks. At the model level, \textbf{HaMo} continually updates a single parameter-efficient adapter using selected trajectories together with historical replay, improving the model's ability to exploit accumulated experience while preserving previously acquired capabilities.

A key design of \textbf{HaMo} is \emph{experience-conditioned adaptation}. Instead of learning only a direct mapping from interaction context to action, the model is additionally conditioned on strategy-level experience retrieved from the harness.
It therefore learns how to interpret and instantiate a retrieved strategy under different observations, objects, states, and interaction histories. The same interface is used during inference: the harness retrieves relevant experience, while the adapted model determines how that experience should be executed in the current environment. In this way, the harness preserves reusable procedural knowledge, whereas model adaptation progressively internalizes the competence required to transfer and execute that knowledge across tasks.

\textbf{HaMo} further addresses the redundancy of continual supervision. Although storing experience externally is inexpensive, repeatedly training on every observed trajectory is unnecessary because many trajectories encode similar strategies. We therefore introduce a \emph{retrieval-coverage-based} selection mechanism that identifies a compact set of representative and complementary trajectories for parameter adaptation. An experience is preferred when it can support many other tasks through retrieval, while the selected set is encouraged to cover distinct strategy contexts. This concentrates parameter updates on broadly reusable procedural knowledge without requiring all stored trajectories to participate in training.

We evaluate \textbf{HaMo} on continual streams of long-horizon tasks across ScienceWorld, ALFWorld, and WebShop, covering both within-domain and cross-domain settings. WebShop provides an interactive Web-agent environment, while the other benchmarks test whether the proposed learning mechanism generalizes beyond a single domain. The results show a clear complementarity between harness- and model-level adaptation: harness-only learning enables rapid reuse of experience but reaches a performance ceiling, whereas model-only adaptation can acquire strong task-specific capabilities but suffers substantial forgetting under distribution shifts. HaMo combines the strengths of both, achieving stronger continual adaptation while better retaining previously acquired capabilities.

\section{Related Work}
\label{sec:related_work}

\subsection{LLMs for Long-Horizon Tasks}
\label{sec:rw_long_horizon}

Long-horizon agents must sustain goal-directed behavior over extended, failure-prone trajectories~\cite{dong2026survey}. Prior work improves reasoning, planning, and web interaction through reasoning-action interleaving~\cite{yao2023react}, stateful benchmarks~\cite{zhou2024webarena}, system-level analyses~\cite{ning2025webagents}, search and reflection~\cite{zhou2024lats}, action knowledge~\cite{zhu2025knowagent}, global planning~\cite{si2026eaglet}, and curriculum-based reinforcement learning~\cite{lai2024autowebglm,kim2026horizon}. Complementary work improves experience reuse via trajectory compression~\cite{yen2025slim}, rollout aggregation~\cite{lee2026aggregation}, global memory~\cite{qian2025memorag}, reusable workflows~\cite{wang2025awm}, and conditional memory~\cite{bi2026memoryneededcharacterizingconditional}. We focus instead on retaining and consolidating reusable experience across continual long-horizon task streams.

\subsection{Continual Learning for LLM Agents}
\label{sec:rw_continual}

External-memory methods retain verbal feedback, skills, workflows, or structured memories without frequent weight updates~\cite{shinn2023reflexion,wang2024voyager,wang2025awm,zhong2024memorybank,gutierrez2025hipporag2,zhang2025memengine}. Parameter-based continual learning instead mitigates forgetting through parameter protection, replay~\cite{sun2020lamol}, continual policy adaptation~\cite{wang2024dialogue}, or multiple retention anchors~\cite{zhang2026compose}. However, neither line directly decides which experience should remain externally retrievable and which should drive parameter updates.

EvolvingAgent~\cite{feng2026evolvingagent} combines evolving experience with continual world-model learning, while Dong et al.~\cite{dong2026survey} highlight the broader connection between external experience and internalized competence. HaMo coordinates these two timescales: rapidly updated strategy-level memory supports immediate reuse, while selected trajectories and replay enable less frequent parameter adaptation.

\begin{figure*}[t]
    \centering
    \includegraphics[width=0.7\linewidth]{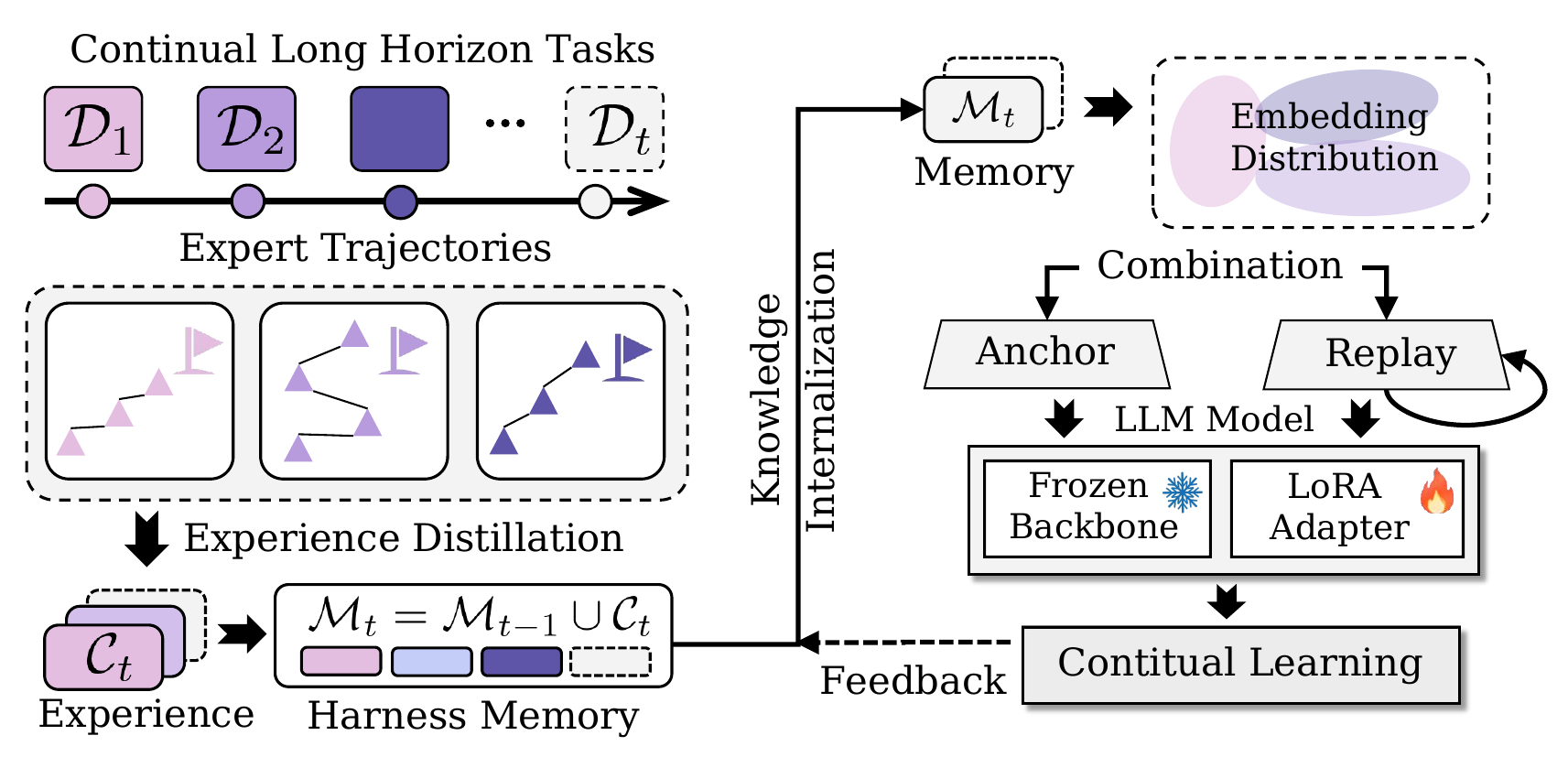}
    \caption{
    Overview of the HaMo framework. 
    HaMo continuously distills expert trajectories into a persistent harness memory, selects representative anchor trajectories based on retrieval coverage, and combines them with historical replay to update a LoRA adapter through experience-conditioned behavioral cloning. An optional feedback step releases memory entries associated with trained high-quality trajectories after adaptation. This collaboration enables fast experience reuse at the harness level and gradual knowledge internalization at the model level.
    }
    \label{fig:method}
    \vspace{-4mm}
\end{figure*}

\section{Method}
\label{sec:method}

HaMo coordinates continual learning through two complementary mechanisms operating at different levels:
a persistent external memory at the harness level and continual parameter adaptation at the model level.
The harness retains reusable strategy-level experience and makes newly acquired knowledge immediately accessible without requiring parameter updates. 
The model, in turn, is continually adapted to interpret retrieved experience, ground it in the current interaction context, and translate it into appropriate actions.
Training and inference share the same initial dialogue format: the user request, the initial observation, and injected matched experience.

\subsection{Harness-Level Memory}
\label{sec:harness-memory}

Long-horizon trajectories contain more than isolated state--action pairs. They often encode reusable plans, intermediate subgoals, and strategies for handling recurring situations. We therefore convert each expert trajectory into a compact strategy-level representation before using it for continual learning. For each trajectory $\tau_i \in D_t$ observed at stage $t$, we use a large language model to extract a solution outline $z_i$. The outline is combined with the corresponding task description $g_i$ and initial observable conditions $o_i^0$ to construct an experience card
\begin{equation}
    c_i = (g_i, o_i^0, z_i)
\end{equation}
The cards extracted from trajectories in $D_t$ form the stage-$t$ collection $C_t$. Each card maintains a one-to-one correspondence with its source trajectory, allowing a selected experience to be mapped back to the trajectory from which it was extracted.

All experience cards observed at the current stage are first added to the external harness:
\begin{equation}
    M_t = M_{t-1} \cup C_t
\end{equation}
where $M_t$ denotes the stage-$t$ memory after adding new cards, and $M_{t-1}$ is the memory retained from the previous stage. This equation describes accumulation; optional post-training release is applied afterward. Persistent storage and trajectory selection serve different purposes in HaMo. The former determines which historical strategies remain externally accessible, whereas the latter determines which trajectories provide supervision for the current parameter update. Consequently, an experience that is not selected for training is not removed from memory and can still be retrieved at later stages.

Although retaining experience in the harness is inexpensive relative to repeated model training, using every newly observed trajectory for parameter adaptation is unnecessary and may introduce substantial redundancy. Multiple trajectories may encode similar strategies and therefore expose the model to nearly identical training contexts. We instead seek a compact subset whose experiences provide broad and complementary coverage over the strategies observed at the current stage.

To estimate the reuse scope of an experience, we embed the textual content of the experience cards and perform leave-one-out retrieval within the candidate collection. For each query card $i$, we retrieve its three nearest neighbors after excluding the card itself, and denote their indices by $\mathcal{N}_3(i)$. We then reverse this retrieval relation. For each candidate card $j$, let
\begin{equation}
    R_j = \{ i \mid j \in \mathcal{N}_3(i) \}
\end{equation}
where $R_j$ contains the query cards for which $j$ is retrieved as a relevant neighbor. Its cardinality,
\begin{equation}
    n_{\mathrm{hits}}(j) = |R_j|
\end{equation}
measures how frequently the strategy represented by card $j$ can potentially support other trajectories.

A high retrieval count alone, however, is insufficient for identifying a useful training subset. Two frequently retrieved cards may support almost the same set of queries, in which case selecting both introduces little additional supervision. We therefore select experiences according to their complementary retrieval coverage. Let $S$ denote the indices of the selected cards. The set of queries covered by $S$ is
\begin{equation}
    F(S) = \bigcup_{j \in S} R_j
\end{equation}
For a new candidate $j$, its additional coverage relative to the current selection is
\begin{equation}
    \Delta(j \mid S)
    =
    R_j \setminus \bigcup_{s \in S} R_s
\end{equation}
Starting from an empty set, we greedily select the candidate with the largest $|\Delta(j \mid S)|$, using lazy priority-queue updates to avoid recomputing the marginal coverage of every candidate at each step. This criterion favors experiences that are broadly reusable while still retaining less frequent experiences when they cover strategy contexts not represented by the current selection.

We continue the selection process until the maximum marginal coverage reaches zero and record the resulting coverage-saturation curve as a function of the number of selected cards. A knee point on this curve determines the stage-specific training budget $N_t^{*}$. The first $N_t^{*}$ selected cards are then mapped back to their corresponding source trajectories, yielding the stage-$t$ anchor set $A_t$. Importantly, this procedure is used only to curate supervision for model adaptation. All experience cards remain available during adaptation, including those whose trajectories are not included in $A_t$. After training, an optional feedback step can release entries associated with trained trajectories, as described in Section~\ref{sec:model-adaptation}.

\subsection{Model-Level Adaptation}
\label{sec:model-adaptation}

The external harness provides persistent access to previously observed strategies, but memory alone does not guarantee that the agent can successfully apply a retrieved strategy to a new interaction. A stored solution outline may have been extracted from a trajectory with different objects, observations, or intermediate states. The model must therefore learn a conditional execution capability: given strategy-level experience, it should ground that experience in the current environment and translate it into actions appropriate for the evolving interaction.

HaMo explicitly learns this capability by conditioning parameter updates on retrieved experience. Let $B_t^{\mathrm{old}}$ denote trajectories sampled from previously encountered tasks for replay. Together with the newly selected anchor trajectories, we construct the stage-$t$ training collection as
\begin{equation}
    B_t = A_t \cup B_t^{\mathrm{old}}
\end{equation}
New anchor trajectories provide supervision for adapting to incoming tasks, while historical trajectories continue to expose the model to behaviors acquired at previous stages.

For every trajectory $\tau_i \in B_t$, let $E_i$ denote the experience retrieved or matched by the harness for that trajectory. The initial dialogue combines the user request $g_i$, initial observation $o_i^0$, and injected experience $E_i$, using the same input format as at test time. Subsequent decisions additionally condition on the evolving interaction history. Let $h_{i,k}$ denote the interaction history preceding the $k$-th action of trajectory $\tau_i$, and let $a_{i,k}$ denote the corresponding expert action. We optimize the experience-conditioned behavioral cloning objective
\begin{equation}
    \mathcal{L}_t(\theta)
    =
    -\frac{1}{|B_t|}
    \sum_{\tau_i \in B_t}
    \sum_k
    \log
    \pi_\theta
    \left(
        a_{i,k}
        \mid
        g_i, o_i^0, h_{i,k}, E_i
    \right)
\end{equation}
For textual actions, the conditional action likelihood is factorized over the corresponding action tokens.

The role of $E_i$ is central to the model-level adaptation. Conventional trajectory supervision primarily learns a direct mapping from the current task context to an action. In contrast, the objective above explicitly trains the model to act under an additional strategy context supplied by the harness. The model is therefore not required to reproduce each stored strategy directly in its parameters. Instead, parameter learning focuses on acquiring a reusable ability to interpret strategy-level guidance and instantiate it under changing objects, observations, states, and interaction histories.

Replay preserves both earlier task behaviors and the ability to execute historical strategies under experience conditioning. The harness retains access to strategies, while replay rehearses their context-dependent use as model parameters evolve.

We implement continual parameter adaptation using a single LoRA adapter while keeping the backbone parameters $W_0$ frozen. At stage $t$, the adapter is initialized from the previous learning stage rather than trained independently:
\begin{equation}
    \theta_t^{\mathrm{init}} = \theta_{t-1}
\end{equation}
Optimizing the experience-conditioned objective on $B_t$ produces
\begin{equation}
    W_t = W_0 + \Delta W(\theta_t)
\end{equation}
where $\Delta W(\theta_t)$ denotes the low-rank update induced by the current adapter. We do not merge the adapter into the backbone or maintain a growing collection of task-specific adapters, so the trainable parameter budget remains fixed throughout the continual learning stream. The harness and the model consequently evolve in different ways: the harness maintains externally accessible experience, whereas the adapter is continually updated to improve and preserve the model's ability to act on that experience.

\paragraph{Optional feedback to the harness.}
After high-quality trajectories have been used for model adaptation, HaMo can optionally release their corresponding memory entries to reclaim storage, implementing the model-to-harness feedback in Figure~\ref{fig:method}. This option is disabled in all reported experiments: all experience cards remain available across stages.

At test time, the initial dialogue likewise combines the user request, initial observation, and injected matched experience. Given $g$ and $o^0$, the harness retrieves experience from the current memory:
\begin{equation}
    E = \operatorname{Retrieve}\bigl((g,o^0), M_t\bigr)
\end{equation}
Unlike the complete-card queries used to estimate retrieval coverage during anchor selection, the inference-time query contains only information available when the new task begins. It therefore does not require an expert trajectory, a ground-truth solution outline, or observations from the future interaction.

The retrieved experience provides strategy-level guidance rather than a fixed action sequence. At interaction step $k$, the policy generates an action according to
\begin{equation}
    a_k
    \sim
    \pi_{\theta_t}
    \left(
        \cdot
        \mid
        g, o^0, h_k, E
    \right)
\end{equation}
where $h_k$ contains the observations and actions accumulated before the current decision. As the environment evolves, newly observed states are incorporated into $h_k$, allowing the model to adapt the retrieved strategy to the actual execution trajectory. The harness therefore preserves a high-level strategy that was useful in previous experience, while the model determines how that strategy should be instantiated under the current objects, conditions, and available actions.

The shared conditioning interface couples learning and inference: the harness preserves reusable strategies, while the model learns to execute them in context. A strategy retained in memory can provide an explicit cue even when later updates weaken the corresponding behavior.

\section{Experimental Setting}
\label{sec:experimental-setting}
\subsection{Datasets and Continual Learning Protocol}
We evaluate HaMo on three interactive benchmarks using their purely text-based environments. \textbf{ScienceWorld}~\cite{wang2022scienceworld} evaluates scientific reasoning through tasks that require agents to navigate an environment, manipulate objects, and carry out experiments. \textbf{ALFWorld}~\cite{shridhar2021alfworld} provides household tasks in a text environment aligned with embodied tasks, requiring agents to locate objects and perform operations such as moving, cleaning, heating, and cooling. \textbf{WebShop}~\cite{yao2022webshop} evaluates instruction-conditioned online shopping: agents search for products, inspect their descriptions, select options, and make purchases that satisfy user requirements. Together, these benchmarks cover scientific experimentation, household manipulation, and web-based decision making, with different action spaces and task semantics.

We consider two continual learning settings. In \emph{within-domain continual learning}, we divide the ScienceWorld training set into four equally sized subsets that are disjoint at the sample level. Every subset covers all task types, so the task-type coverage remains unchanged across stages even though the training samples are different. We introduce one subset at each successive stage, simulating within-domain data-incremental continual learning over a fixed set of task types rather than the sequential introduction of new task types. After every stage, we evaluate the agent on the entire ScienceWorld test set, using the same evaluation set throughout the stream. In \emph{cross-domain continual learning}, the agent learns sequentially on ScienceWorld (S), ALFWorld (A), and WebShop (W). After each stage, we evaluate all domains encountered so far, allowing us to examine both adaptation to the current domain and retention of earlier capabilities. The main cross-domain experiment uses the order S $\to$ A $\to$ W; the order ablation additionally considers A $\to$ W $\to$ S and W $\to$ S $\to$ A.

\subsection{Baselines}
We use Qwen3~\cite{qwen3technicalreport} as the common backbone, evaluating Qwen3-8B and Qwen3-14B in the within-domain setting. Cross-domain experiments and component and training-order ablations use Qwen3-8B, while the model-scale ablation additionally evaluates Qwen3-1.7B and Qwen3-4B. Thinking mode is disabled for all Qwen3 agents. \textbf{Base} denotes the backbone without the additional agent training or external harness considered here. \textbf{Base+Harness} augments the backbone with external experience while keeping its parameters fixed, isolating adaptation through the harness. We also compare two agent-training approaches instantiated on the Qwen3 backbone. \textbf{AgentTuning}~\cite{zeng2024agenttuning} develops agent capabilities through instruction tuning on interaction trajectories; its original formulation mixes agent trajectories with general instruction data. \textbf{ETO}~\cite{song2024eto}, or Exploration-Based Trajectory Optimization, uses exploration to collect failed trajectories and constructs trajectory preference pairs for policy optimization. These comparisons evaluate how agent-training approaches behave under our continual learning settings.

\textbf{HaMo} combines a persistent strategy memory, model adaptation conditioned on experience, and historical replay, as described in Section~\ref{sec:method}. Its model updates use a single continually updated LoRA adapter with a frozen backbone. In the component ablation, \textbf{Harness only} corresponds to Base+Harness, while \textbf{Model only} performs sequential parameter adaptation without either the external harness or historical replay. This comparison examines the contribution of coordinating external experience with model adaptation.
Implementation settings and framework pseudocode are provided in Appendices~\ref{app:algorithm} and~\ref{app:settings}, respectively.

\begin{figure*}[!tp]
    \centering
    \begin{subfigure}[t]{0.24\textwidth}
        \centering
        \includegraphics[width=\linewidth]{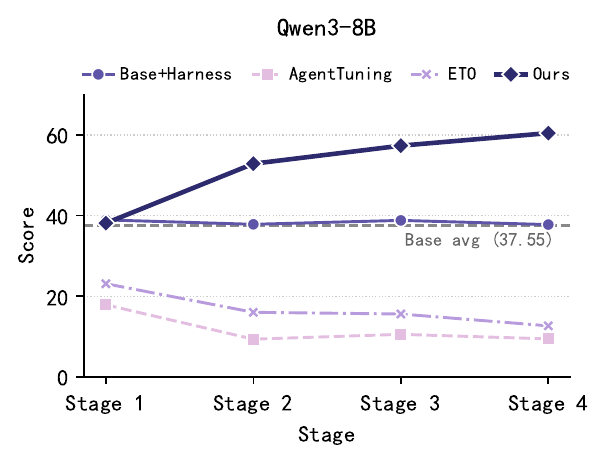}
        \caption{Performance(Qwen3-8b)}
        \label{fig:sci_stage_8b}
    \end{subfigure}\hfill
    \begin{subfigure}[t]{0.24\textwidth}
        \centering
        \includegraphics[width=\linewidth]{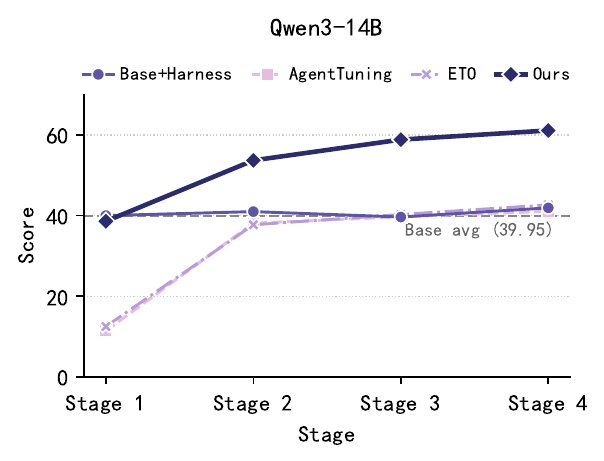}
        \caption{Performance(Qwen3-14b)}
        \label{fig:sci_stage_14b}
    \end{subfigure}\hfill
    \begin{subfigure}[t]{0.24\textwidth}
        \centering
        \includegraphics[width=\linewidth]{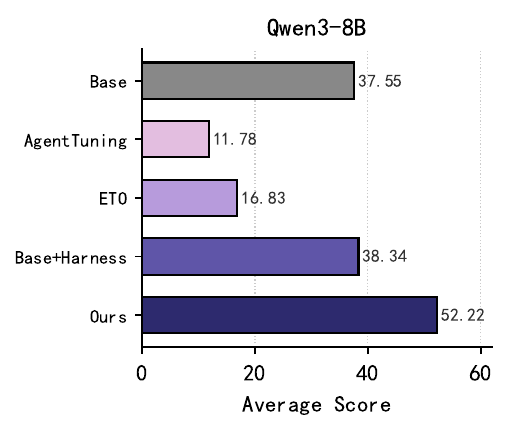}
        \caption{Performance(Qwen3-8b)}
        \label{fig:sci_avg_8b}
    \end{subfigure}\hfill
    \begin{subfigure}[t]{0.24\textwidth}
        \centering
        \includegraphics[width=\linewidth]{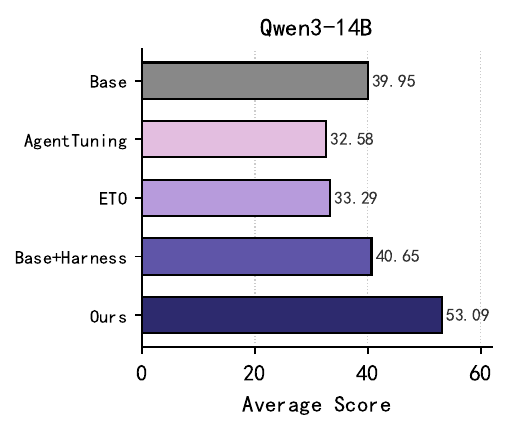}
        \caption{Performance(Qwen3-14b)}
        \label{fig:sci_avg_14b}
    \end{subfigure}
    \caption{ 
    Within-domain data-incremental continual learning performance (stage-wise and average performance) on ScienceWorld. 
    The training set is divided into four equally sized, sample-disjoint subsets, each covering all task types.
    Each stage introduces one subset and evaluates the entire test set.
    Average reports the mean stage score for adapting methods and the fixed reference score for Base. Higher is better.
    }
    \label{fig:scienceworld_performance}
    \Description{Four panels show stage-wise performance for Qwen3-8B and Qwen3-14B on the left, and average performance for the same models on the right.}
    \vspace{-4mm}
\end{figure*}

\begin{figure*}[t]
    \centering
    \includegraphics[width=1.0\linewidth]{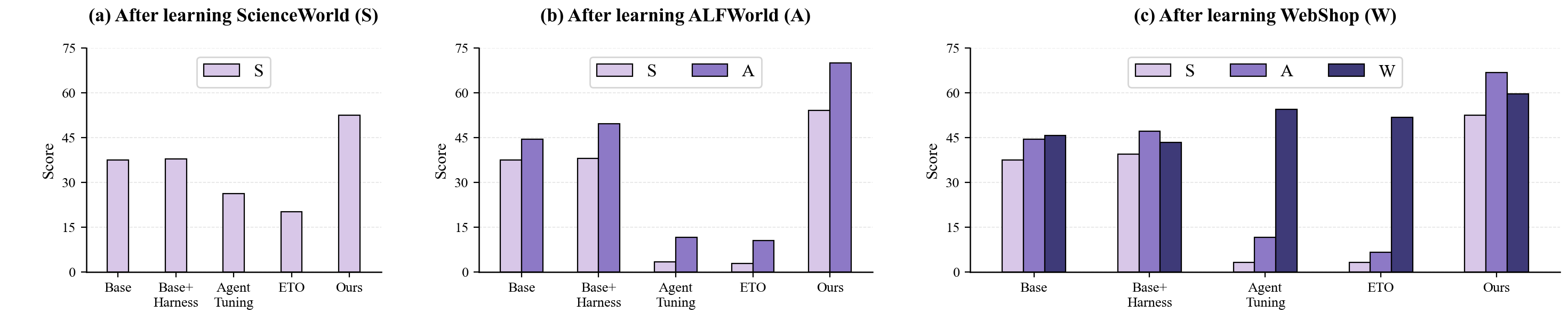}
    \caption{
    Cross-domain continual learning with Qwen3-8B in the order ScienceWorld (S), ALFWorld (A), and WebShop (W). Panels (a)--(c) show performance after each training stage. Bar groups compare methods, and colors indicate evaluation domains. Higher scores are better.
    }
    \label{fig:cross_domain}
    \vspace{-4mm}
\end{figure*}

\subsection{Metrics}
We use the native scoring mechanism of each text environment as the performance metric. For ScienceWorld, we report the environment's task-progress score, which credits intermediate progress toward completing the scientific task. For ALFWorld, we report the task-completion score, averaged across episodes as a success rate. For WebShop, we report the environment reward measuring how well the purchased product satisfies the requested specifications, retaining partial credit rather than reducing the outcome to binary success. Scores are presented on a 0--100 scale, with higher values indicating better performance. Each evaluation episode is limited to \textbf{50 agent--environment interaction turns} and ends earlier if the environment terminates the task.

For the within-domain experiment, \emph{Average} summarizes the four reported stage scores; the Base score gives the fixed backbone reference. In the cross-domain performance plots, each panel corresponds to a completed training stage and shows scores on the evaluated domains. We assess retention by comparing a domain's score immediately after learning it with its scores after subsequent stages. The training-order ablation also reports the sum of the three final-domain scores as a descriptive aggregate. Since the benchmarks use different scoring semantics, this sum complements the per-domain results rather than representing a pooled task-success rate.

\section{Results}
\label{sec:results}
\subsection{Main Results}
\subsubsection{Within-Domain Continual Learning}

Figure~\ref{fig:scienceworld_performance} shows that HaMo improves on the full ScienceWorld test set as additional training subsets arrive, for both model sizes. With Qwen3-8B, its score increases from 38.13 to 60.49, yielding a four-stage average of 52.22. With Qwen3-14B, it increases from 38.61 to 61.15, with an average of 53.09. These averages exceed Base+Harness by 13.88 and 12.44 points, respectively, using the reported values. The final-stage gains over the unadapted backbones are 22.94 points for 8B and 21.20 points for 14B.

The advantage emerges over successive stages rather than from uniformly stronger initial performance. At Stage~1, HaMo scores slightly below Base+Harness for both backbones, but it leads all listed methods from Stage~2 onward. Base+Harness remains close to the backbone reference throughout the stream, suggesting that external experience alone provides limited improvement in this setting. AgentTuning and ETO exhibit different trajectories across model sizes: their 8B scores decline overall, whereas their 14B scores recover after a low first stage. Nevertheless, their four-stage averages remain below those of HaMo. These results support the value of jointly adapting the model and its experience context for sustained improvement within ScienceWorld; the stage averages alone do not isolate retention of individual task types.

\begin{figure*}[t]
    \centering
    \includegraphics[width=0.95\linewidth]{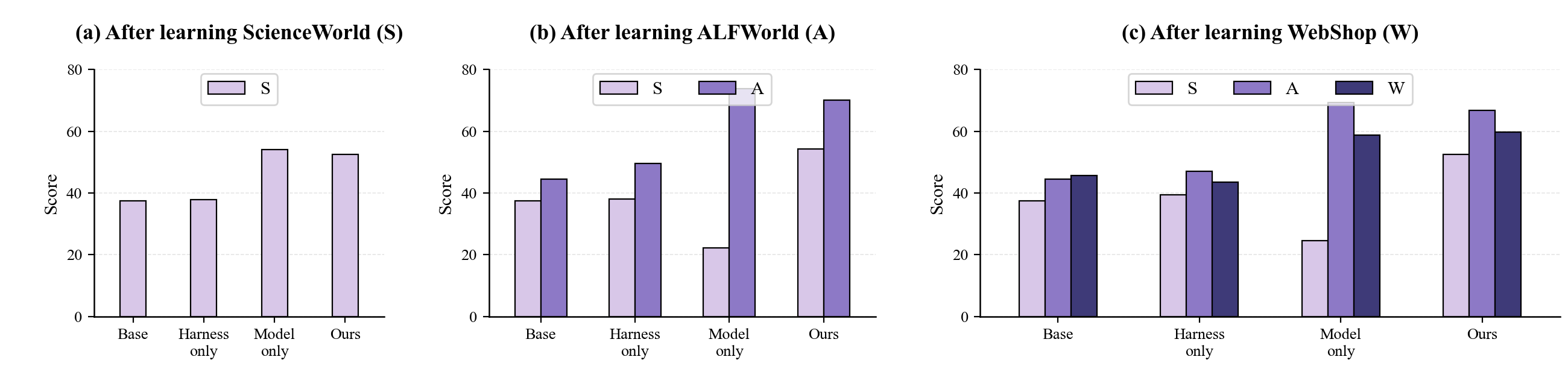}
    \caption{
    Component ablation on Qwen3-8B with the training order ScienceWorld (S), ALFWorld (A), and WebShop (W). Each group reports performance after training on the indicated dataset. 
    }
    \label{fig:component_ablation}
    \vspace{-4mm}
\end{figure*}

\begin{figure*}[t]
    \centering
    \includegraphics[width=0.95\linewidth]{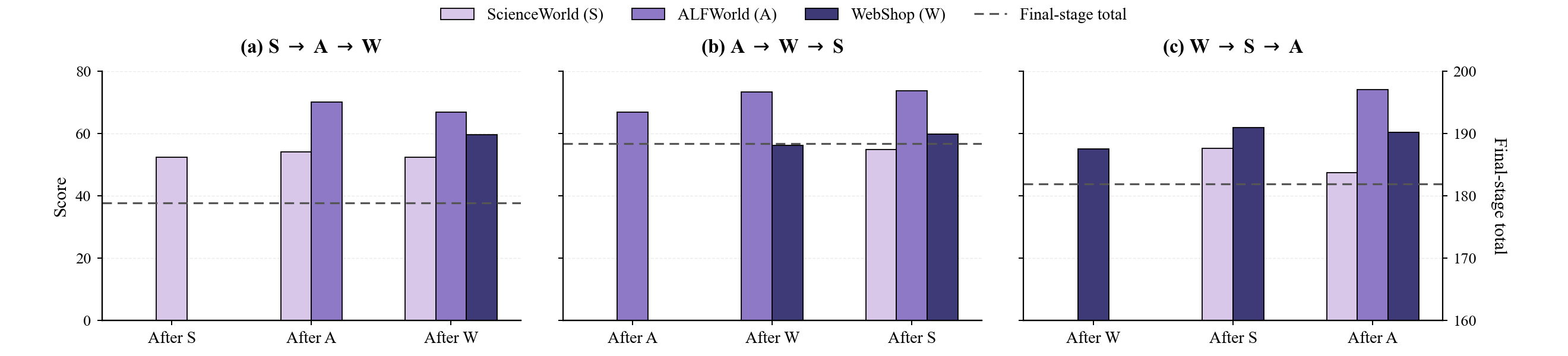}
    \caption{
    Training-order ablation on Qwen3-8B. S, A, and W denote ScienceWorld, ALFWorld, and WebShop, respectively. Final-stage total (S+A+W) sums the three dataset scores measured after the final training stage for each training order.
    }
    \label{fig:order_ablation}
    \vspace{-4mm}
\end{figure*}

\subsubsection{Cross-Domain Continual Learning}
Figure~\ref{fig:cross_domain} evaluates the S $\to$ A $\to$ W stream. After the final stage, HaMo obtains 52.43 on ScienceWorld, 66.79 on ALFWorld, and 59.64 on WebShop, outperforming every listed baseline on each domain. Relative to Base, these are gains of 14.88, 22.26, and 13.96 points. Relative to the strongest non-HaMo baseline for each final-domain score, the gains are 13.02 points on ScienceWorld, 19.71 on ALFWorld, and 5.16 on WebShop.

The intermediate evaluations reveal how these gains arise. HaMo scores 52.43 on ScienceWorld after the first stage, improves to 54.19 after learning ALFWorld, and returns to 52.43 after learning WebShop. Thus, its final ScienceWorld score matches its score immediately after ScienceWorld training, although it is 1.76 points below the intermediate peak. Its ALFWorld score decreases moderately from 70.08 to 66.79 after WebShop training. In contrast, AgentTuning's ScienceWorld score falls from 26.24 to 3.13, and ETO's falls from 20.16 to 3.20. Both methods improve over Base on the final WebShop task but finish below Base on the two earlier domains. Under this protocol, successful adaptation to the latest domain therefore does not ensure retention of earlier capabilities.

Base+Harness largely preserves the backbone's earlier-domain performance, but its final WebShop score of 43.42 is below the Base score of 45.68. HaMo combines stronger acquisition of new-domain capabilities with substantially better earlier-domain performance than the trajectory-training baselines. This pattern is consistent with the intended complementarity of persistent strategy access and continual learning of how to execute those strategies.

\subsection{Ablation Results}

\subsubsection{Harness--Model Collaboration}

Figure~\ref{fig:component_ablation} compares HaMo with harness-only and model-only adaptation. Harness only provides relatively small changes from Base, whereas Model only substantially improves newly learned tasks: it obtains 53.98 on ScienceWorld after the first stage and 73.72 on ALFWorld after the second. However, its ScienceWorld score drops to 22.18 after ALFWorld training and finishes at 24.57, a decline of 29.41 points from its first-stage score. HaMo instead finishes at 52.43 on ScienceWorld, exceeding Model only by 27.86 points and matching its own first-stage score.

This retention benefit comes with a measurable tradeoff. Model only achieves a higher final ALFWorld score than HaMo (69.34 versus 66.79), while HaMo is slightly better on WebShop (59.64 versus 58.74). HaMo therefore does not dominate parameter-only adaptation on every domain. Its principal advantage in this ablation is a more favorable balance between acquisition and retention, particularly on the earliest domain. Because Model only omits both the external harness and historical replay, this comparison supports the complete HaMo design rather than isolating the effect of external memory. It does not separately identify the contributions of anchor selection, replay, and experience conditioning.

\subsubsection{Sensitivity to Training Order}


\begin{figure}
    \centering
    \includegraphics[width=0.7\linewidth]{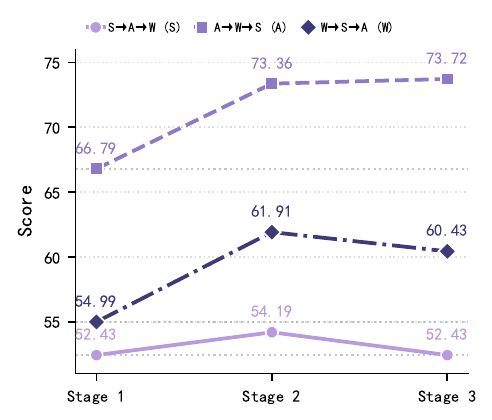}
    \caption{
    Performance on the first-trained domain throughout continual learning under three training orders. The curves track ScienceWorld in S → A → W, ALFWorld in A → W → S, and WebShop in W → S → A.
     }
    \label{fig:first_dataset_performance}
    \vspace{-4mm}
\end{figure}

\begin{table*}[t]
\centering
\caption{WebShop case study. Trajectories are summarized from the recorded interactions. Purple text backgrounds mark HaMo's key constraint-grounding behaviors. Scores are reported on the same 0--100 scale as the main experiments.}
\label{tab:webshop-case}
\small
\setlength{\tabcolsep}{5pt}
\setlength{\fboxsep}{1.5pt}
\renewcommand{\arraystretch}{1.15}

\begin{tabular}{p{0.13\textwidth} p{0.40\textwidth} p{0.40\textwidth}}
\toprule
\textbf{Stage} & \textbf{HaMo} & \textbf{Base} \\
\midrule

User Goal
& Chocolate-coated wafers; caramel flavor; 4.41 ounce; price below \$60.
& Chocolate-coated wafers; caramel flavor; 4.41 ounce; price below \$60. \\

\midrule

Search
& Searches for ``chocolate coated wafers 4.41 ounce''.
& Searches for ``caramel chocolate coated wafers 4.41 ounce''. \\

Candidate
& Opens Loacker Tortina Chocolate Coated Wafer (\$41.99). Notices ``Triple Dark'' in the title and \hlpurple{checks the available flavor options}.
& Opens Loacker Quadratini Hazelnut Wafer Cookies (\$14.99), incorrectly treating the hazelnut product as a caramel match. \\

Verification
& Attempts to select the requested options: \hlpurple{\texttt{click[caramel]}} followed by \hlpurple{\texttt{click[4.41 ounce]}}; neither changes the visible observation.
& The product page offers no caramel option. Makes no flavor or size selection before buying. \\

Purchase
& \hlpurple{Checks \$41.99 $<$ \$60} and executes \texttt{click[buy now]} after attempting both option selections.
& Executes \texttt{click[buy now]} immediately after opening the mismatched product. \\

Interactions
& 5 actions; completes the purchase.
& 3 actions; completes the purchase. \\

\midrule

Score
& \textbf{100}
& 25 \\

\bottomrule
\end{tabular}

\end{table*}

\begin{figure*}[t]
    \centering
    \includegraphics[width=0.95\linewidth]{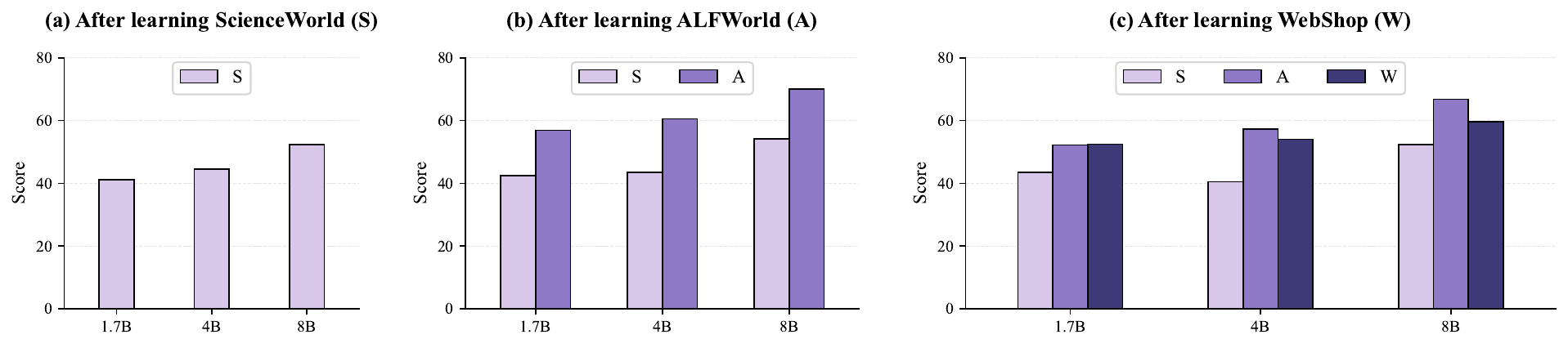}
    \caption{
    Performance of HaMo across model scales (1.7B, 4B, and 8B) with the training order ScienceWorld (S), ALFWorld (A), and WebShop (W). 
    The largest LLM backbone achieves the strongest overall results, supporting HaMo's applicability across model capacities.
     }
    \label{fig:scale_abla}
    \vspace{-4mm}
\end{figure*}


Figure~\ref{fig:order_ablation} examines three cyclic training orders. HaMo achieves final-stage totals of 178.86 for S $\to$ A $\to$ W, 188.41 for A $\to$ W $\to$ S, and 181.94 for W $\to$ S $\to$ A. All final-domain scores exceed the corresponding Base references in the main cross-domain experiment, although their distribution depends on training order. Among the tested orders, A $\to$ W $\to$ S produces the highest total, while W $\to$ S $\to$ A achieves the highest final ALFWorld and WebShop scores (74.09 and 60.43).

Figure~\ref{fig:first_dataset_performance} tracks retention on the first-trained domain. Final ScienceWorld performance matches its initial 52.43 in S $\to$ A $\to$ W; ALFWorld improves from 66.79 to 73.72 in A $\to$ W $\to$ S, and WebShop from 54.99 to 60.43 in W $\to$ S $\to$ A. Retention remains order-dependent: in the last sequence, ScienceWorld falls from 55.19 to 47.42 after subsequent adaptation, showing that forgetting is mitigated but not eliminated.

\subsubsection{Robustness to Model Scale}

Figure ~\ref{fig:scale_abla} evaluates HaMo with Qwen3 models of different scales, including 1.7B, 4B, and 8B parameters. HaMo remains effective across all three model sizes throughout the S → A → W continual learning stream, indicating that its benefit is not restricted to a particular model capacity. Larger LLMs generally show stronger acquisition of newly introduced tasks, especially after adapting to later domains, suggesting that increased model capacity provides greater flexibility for incorporating new procedural knowledge.


\subsubsection{Case Study: Experience-Guided Constraint Verification}
\label{sec:case-study}
Table~\ref{tab:webshop-case} illustrates experience-guided constraint verification. The task requires caramel-flavored, chocolate-coated wafers in a 4.41-ounce size below \$60. Base purchases a hazelnut product without resolving the flavor mismatch, scoring 25. HaMo recognizes that the candidate's ``Triple Dark'' title does not establish a caramel match and checks the available options, scoring 100. This follows the retrieved card's instruction to recheck requirements against the current task; HaMo also checks the current \$60 budget rather than the reference's \$70 limit. The actions \texttt{click[caramel]} and \texttt{click[4.41 ounce]} produce neither a visible observation change nor an immediate reward. They indicate attempted grounding of the requested attributes, but do not establish state changes or causally explain the reward gain. The case illustrates attention to option availability and current constraints before purchase, without isolating the contributions of retrieved guidance and parameter adaptation.


\section{Conclusion}
We presented HaMo, a harness–model collaborative framework for continual learning in long-horizon agents. HaMo combines persistent strategy-level memory with experience-conditioned parameter adaptation, enabling newly acquired procedural knowledge to be reused immediately while gradually improving the model’s ability to interpret and execute retrieved strategies. Experiments on ScienceWorld, ALFWorld, and WebShop show that HaMo achieves a better balance between adaptation and retention than harness-only or model-only approaches, with stronger robustness under both within-domain and cross-domain continual learning. These results highlight the value of coordinating external experience accumulation with internal model adaptation, and suggest a promising direction for building long-horizon agents that can continually learn from interaction while preserving previously acquired capabilities.

\section*{Limitations}
\label{app:limitations}

HaMo can offset part of the forgetting induced by cross-domain training, but it does not fully prevent catastrophic forgetting. Subsequent adaptation may still reduce performance on previously learned domains, and the degree of retention depends on the domain sequence. The current results therefore support partial mitigation of forgetting rather than complete immunity to it.

\section*{Future Work}
\label{app:future}

\paragraph{Adaptive experience use.}
A useful extension is to learn when to retrieve a strategy, when to reject an unsuitable card, and when to refresh guidance during execution. Such decisions could combine retrieval confidence with observed task progress and explicit checks of strategy preconditions. This would allow the agent to respond to misleading experience rather than treating retrieval similarity as sufficient evidence of applicability.

\paragraph{Stronger evaluation and broader environments.}
Controlled ablations of replay, anchor selection, experience conditioning, and decoding temperature would clarify their individual contributions and interactions. Repeated runs and longer task streams could assess variability and persistent forgetting, while end-to-end cost measurements would expose the tradeoff between performance and memory maintenance. Extending the framework to multimodal agents and changing web environments would further test whether experience-conditioned execution transfers beyond the present text-based benchmarks.



\bibliographystyle{ACM-Reference-Format}
\bibliography{ref}


\begin{thebibliography}{41}


\ifx \showCODEN    \undefined \def \showCODEN     #1{\unskip}     \fi
\ifx \showISBNx    \undefined \def \showISBNx     #1{\unskip}     \fi
\ifx \showISBNxiii \undefined \def \showISBNxiii  #1{\unskip}     \fi
\ifx \showISSN     \undefined \def \showISSN      #1{\unskip}     \fi
\ifx \showLCCN     \undefined \def \showLCCN      #1{\unskip}     \fi
\ifx \shownote     \undefined \def \shownote      #1{#1}          \fi
\ifx \showarticletitle \undefined \def \showarticletitle #1{#1}   \fi
\ifx \showURL      \undefined \def \showURL       {\relax}        \fi
\providecommand\bibfield[2]{#2}
\providecommand\bibinfo[2]{#2}
\providecommand\natexlab[1]{#1}
\providecommand\showeprint[2][]{arXiv:#2}

\bibitem[Bi et~al\mbox{.}(2026)]%
        {bi2026memoryneededcharacterizingconditional}
\bibfield{author}{\bibinfo{person}{Zhen Bi}, \bibinfo{person}{Xueshu Chen}, \bibinfo{person}{Yan Wang}, \bibinfo{person}{Zhizhi Peng}, \bibinfo{person}{Haosen Hong}, \bibinfo{person}{Zhen Wang}, \bibinfo{person}{Zhixuan Chu}, \bibinfo{person}{Bingyu Zhu}, {and} \bibinfo{person}{Jungang Lou}.} \bibinfo{year}{2026}\natexlab{}.
\newblock \bibinfo{title}{Memory Is Not Always Needed: Characterizing Conditional Memory in Scientific Reasoning}.
\newblock
\showeprint[arxiv]{2608.23982}~[cs.AI]
\urldef\tempurl%
\url{https://arxiv.org/abs/2608.23982}
\showURL{%
\tempurl}


\bibitem[Chen et~al\mbox{.}(2023)]%
        {chen2023fireact}
\bibfield{author}{\bibinfo{person}{Baian Chen}, \bibinfo{person}{Chang Shu}, \bibinfo{person}{Ehsan Shareghi}, \bibinfo{person}{Nigel Collier}, \bibinfo{person}{Karthik Narasimhan}, {and} \bibinfo{person}{Shunyu Yao}.} \bibinfo{year}{2023}\natexlab{}.
\newblock \showarticletitle{Fireact: Toward language agent fine-tuning}.
\newblock \bibinfo{journal}{\emph{arXiv preprint arXiv:2310.05915}} (\bibinfo{year}{2023}).
\newblock


\bibitem[Deng et~al\mbox{.}(2023)]%
        {deng2023mind2web}
\bibfield{author}{\bibinfo{person}{Xiang Deng}, \bibinfo{person}{Yu Gu}, \bibinfo{person}{Boyuan Zheng}, \bibinfo{person}{Shijie Chen}, \bibinfo{person}{Sam Stevens}, \bibinfo{person}{Boshi Wang}, \bibinfo{person}{Huan Sun}, {and} \bibinfo{person}{Yu Su}.} \bibinfo{year}{2023}\natexlab{}.
\newblock \showarticletitle{Mind2web: Towards a generalist agent for the web}.
\newblock \bibinfo{journal}{\emph{Advances in Neural Information Processing Systems}}  \bibinfo{volume}{36} (\bibinfo{year}{2023}), \bibinfo{pages}{28091--28114}.
\newblock


\bibitem[Dong et~al\mbox{.}(2026)]%
        {dong2026survey}
\bibfield{author}{\bibinfo{person}{Guanting Dong}, \bibinfo{person}{Xiaoshuai Song}, \bibinfo{person}{Yuyang Hu}, \bibinfo{person}{Jiajie Jin}, \bibinfo{person}{Chenghao Zhang}, \bibinfo{person}{Yifei Chen}, \bibinfo{person}{Xiaoxi Li}, \bibinfo{person}{Huaying Yuan}, \bibinfo{person}{Xinyu Yang}, \bibinfo{person}{Tongyu Wen}, {et~al\mbox{.}}} \bibinfo{year}{2026}\natexlab{}.
\newblock \showarticletitle{Towards Long-Horizon Agents: A Survey}.
\newblock  (\bibinfo{year}{2026}).
\newblock


\bibitem[Drouin et~al\mbox{.}(2024)]%
        {drouin2024workarena}
\bibfield{author}{\bibinfo{person}{Alexandre Drouin}, \bibinfo{person}{Maxime Gasse}, \bibinfo{person}{Massimo Caccia}, \bibinfo{person}{Issam~H Laradji}, \bibinfo{person}{Manuel Del~Verme}, \bibinfo{person}{Tom Marty}, \bibinfo{person}{L{\'e}o Boisvert}, \bibinfo{person}{Megh Thakkar}, \bibinfo{person}{Quentin Cappart}, \bibinfo{person}{David Vazquez}, {et~al\mbox{.}}} \bibinfo{year}{2024}\natexlab{}.
\newblock \showarticletitle{WorkArena: How capable are web agents at solving common knowledge work tasks?}
\newblock \bibinfo{journal}{\emph{arXiv preprint arXiv:2403.07718}} (\bibinfo{year}{2024}).
\newblock


\bibitem[Feng et~al\mbox{.}(2025)]%
        {feng2026evolvingagent}
\bibfield{author}{\bibinfo{person}{Tongtong Feng}, \bibinfo{person}{Xin Wang}, \bibinfo{person}{Zekai Zhou}, \bibinfo{person}{Ren Wang}, \bibinfo{person}{Yuwei Zhan}, \bibinfo{person}{Guangyao Li}, \bibinfo{person}{Qing Li}, {and} \bibinfo{person}{Wenwu Zhu}.} \bibinfo{year}{2025}\natexlab{}.
\newblock \showarticletitle{EvolvingAgent: Curriculum Self-evolving Agent with Continual World Model for Long-Horizon Tasks}.
\newblock \bibinfo{journal}{\emph{arXiv preprint arXiv:2502.05907}} (\bibinfo{year}{2025}).
\newblock


\bibitem[Guti{\'e}rrez et~al\mbox{.}(2025)]%
        {gutierrez2025hipporag2}
\bibfield{author}{\bibinfo{person}{Bernal~Jim{\'e}nez Guti{\'e}rrez}, \bibinfo{person}{Yiheng Shu}, \bibinfo{person}{Weijian Qi}, \bibinfo{person}{Sizhe Zhou}, {and} \bibinfo{person}{Yu Su}.} \bibinfo{year}{2025}\natexlab{}.
\newblock \showarticletitle{From {RAG} to Memory: Non-Parametric Continual Learning for Large Language Models}. In \bibinfo{booktitle}{\emph{Forty-second International Conference on Machine Learning}}.
\newblock
\urldef\tempurl%
\url{https://openreview.net/forum?id=LWH8yn4HS2}
\showURL{%
\tempurl}


\bibitem[He et~al\mbox{.}(2024)]%
        {he2024webvoyager}
\bibfield{author}{\bibinfo{person}{Hongliang He}, \bibinfo{person}{Wenlin Yao}, \bibinfo{person}{Kaixin Ma}, \bibinfo{person}{Wenhao Yu}, \bibinfo{person}{Yong Dai}, \bibinfo{person}{Hongming Zhang}, \bibinfo{person}{Zhenzhong Lan}, {and} \bibinfo{person}{Dong Yu}.} \bibinfo{year}{2024}\natexlab{}.
\newblock \showarticletitle{Webvoyager: Building an end-to-end web agent with large multimodal models}. In \bibinfo{booktitle}{\emph{Proceedings of the 62nd Annual Meeting of the Association for Computational Linguistics (Volume 1: Long Papers)}}. \bibinfo{pages}{6864--6890}.
\newblock


\bibitem[Kim et~al\mbox{.}(2026)]%
        {kim2026horizon}
\bibfield{author}{\bibinfo{person}{Sunghwan Kim}, \bibinfo{person}{Junhee Cho}, \bibinfo{person}{Beong-Woo Kwak}, \bibinfo{person}{Taeyoon Kwon}, \bibinfo{person}{Liang Wang}, \bibinfo{person}{Nan Yang}, \bibinfo{person}{Xingxing Zhang}, \bibinfo{person}{Furu Wei}, {and} \bibinfo{person}{Jinyoung Yeo}.} \bibinfo{year}{2026}\natexlab{}.
\newblock \showarticletitle{On Training Large Language Models for Long-Horizon Tasks: An Empirical Study of Horizon Length}. In \bibinfo{booktitle}{\emph{Proceedings of the 43rd International Conference on Machine Learning}} \emph{(\bibinfo{series}{Proceedings of Machine Learning Research}, Vol.~\bibinfo{volume}{306})}, \bibfield{editor}{\bibinfo{person}{Tong Zhang}, \bibinfo{person}{Miroslav Dudik}, \bibinfo{person}{Martin Jaggi}, \bibinfo{person}{Alekh Agarwal}, \bibinfo{person}{Sharon Li}, \bibinfo{person}{Dale Schuurmans}, \bibinfo{person}{Jerry Zhu}, \bibinfo{person}{Felix Berkenkamp}, \bibinfo{person}{Hanze Dong}, {and} \bibinfo{person}{Alberto Bietti}} (Eds.). \bibinfo{publisher}{PMLR}, \bibinfo{pages}{57309--57337}.
\newblock
\urldef\tempurl%
\url{https://proceedings.mlr.press/v306/kim26h.html}
\showURL{%
\tempurl}


\bibitem[Lai et~al\mbox{.}(2024)]%
        {lai2024autowebglm}
\bibfield{author}{\bibinfo{person}{Hanyu Lai}, \bibinfo{person}{Xiao Liu}, \bibinfo{person}{Iat~Long Iong}, \bibinfo{person}{Shuntian Yao}, \bibinfo{person}{Yuxuan Chen}, \bibinfo{person}{Pengbo Shen}, \bibinfo{person}{Hao Yu}, \bibinfo{person}{Hanchen Zhang}, \bibinfo{person}{Xiaohan Zhang}, \bibinfo{person}{Yuxiao Dong}, {et~al\mbox{.}}} \bibinfo{year}{2024}\natexlab{}.
\newblock \showarticletitle{Autowebglm: A large language model-based web navigating agent}. In \bibinfo{booktitle}{\emph{Proceedings of the 30th ACM SIGKDD Conference on Knowledge Discovery and Data Mining}}. \bibinfo{pages}{5295--5306}.
\newblock


\bibitem[Lee et~al\mbox{.}(2026)]%
        {lee2026aggregation}
\bibfield{author}{\bibinfo{person}{Yoonsang Lee}, \bibinfo{person}{Howard Yen}, \bibinfo{person}{Xi Ye}, {and} \bibinfo{person}{Danqi Chen}.} \bibinfo{year}{2026}\natexlab{}.
\newblock \showarticletitle{Agentic aggregation for parallel scaling of long-horizon agentic tasks}.
\newblock \bibinfo{journal}{\emph{arXiv preprint arXiv:2604.11753}} (\bibinfo{year}{2026}).
\newblock


\bibitem[Luo et~al\mbox{.}(2025)]%
        {luo2023forgetting}
\bibfield{author}{\bibinfo{person}{Yun Luo}, \bibinfo{person}{Zhen Yang}, \bibinfo{person}{Fandong Meng}, \bibinfo{person}{Yafu Li}, \bibinfo{person}{Jie Zhou}, {and} \bibinfo{person}{Yue Zhang}.} \bibinfo{year}{2025}\natexlab{}.
\newblock \showarticletitle{An empirical study of catastrophic forgetting in large language models during continual fine-tuning}.
\newblock \bibinfo{journal}{\emph{IEEE Transactions on Audio, Speech and Language Processing}} (\bibinfo{year}{2025}).
\newblock


\bibitem[Majumder et~al\mbox{.}(2023)]%
        {majumder2023clin}
\bibfield{author}{\bibinfo{person}{Bodhisattwa~Prasad Majumder}, \bibinfo{person}{Bhavana~Dalvi Mishra}, \bibinfo{person}{Peter Jansen}, \bibinfo{person}{Oyvind Tafjord}, \bibinfo{person}{Niket Tandon}, \bibinfo{person}{Li Zhang}, \bibinfo{person}{Chris Callison-Burch}, {and} \bibinfo{person}{Peter Clark}.} \bibinfo{year}{2023}\natexlab{}.
\newblock \showarticletitle{Clin: A continually learning language agent for rapid task adaptation and generalization}.
\newblock \bibinfo{journal}{\emph{arXiv preprint arXiv:2310.10134}} (\bibinfo{year}{2023}).
\newblock


\bibitem[Ning et~al\mbox{.}(2025)]%
        {ning2025webagents}
\bibfield{author}{\bibinfo{person}{Liangbo Ning}, \bibinfo{person}{Ziran Liang}, \bibinfo{person}{Zhuohang Jiang}, \bibinfo{person}{Haohao Qu}, \bibinfo{person}{Yujuan Ding}, \bibinfo{person}{Wenqi Fan}, \bibinfo{person}{Xiao-yong Wei}, \bibinfo{person}{Shanru Lin}, \bibinfo{person}{Hui Liu}, \bibinfo{person}{Philip~S Yu}, {et~al\mbox{.}}} \bibinfo{year}{2025}\natexlab{}.
\newblock \showarticletitle{A survey of webagents: Towards next-generation ai agents for web automation with large foundation models}. In \bibinfo{booktitle}{\emph{Proceedings of the 31st ACM SIGKDD Conference on Knowledge Discovery and Data Mining V. 2}}. \bibinfo{pages}{6140--6150}.
\newblock


\bibitem[Park et~al\mbox{.}(2023)]%
        {park2023generative}
\bibfield{author}{\bibinfo{person}{Joon~Sung Park}, \bibinfo{person}{Joseph O'Brien}, \bibinfo{person}{Carrie~Jun Cai}, \bibinfo{person}{Meredith~Ringel Morris}, \bibinfo{person}{Percy Liang}, {and} \bibinfo{person}{Michael~S Bernstein}.} \bibinfo{year}{2023}\natexlab{}.
\newblock \showarticletitle{Generative agents: Interactive simulacra of human behavior}. In \bibinfo{booktitle}{\emph{Proceedings of the 36th annual acm symposium on user interface software and technology}}. \bibinfo{pages}{1--22}.
\newblock


\bibitem[Putta et~al\mbox{.}(2024)]%
        {putta2024agentq}
\bibfield{author}{\bibinfo{person}{Pranav Putta}, \bibinfo{person}{Edmund Mills}, \bibinfo{person}{Naman Garg}, \bibinfo{person}{Sumeet Motwani}, \bibinfo{person}{Chelsea Finn}, \bibinfo{person}{Divyansh Garg}, {and} \bibinfo{person}{Rafael Rafailov}.} \bibinfo{year}{2024}\natexlab{}.
\newblock \showarticletitle{Agent q: Advanced reasoning and learning for autonomous ai agents}.
\newblock \bibinfo{journal}{\emph{arXiv preprint arXiv:2408.07199}} (\bibinfo{year}{2024}).
\newblock


\bibitem[Qian et~al\mbox{.}(2025)]%
        {qian2025memorag}
\bibfield{author}{\bibinfo{person}{Hongjin Qian}, \bibinfo{person}{Zheng Liu}, \bibinfo{person}{Peitian Zhang}, \bibinfo{person}{Kelong Mao}, \bibinfo{person}{Defu Lian}, \bibinfo{person}{Zhicheng Dou}, {and} \bibinfo{person}{Tiejun Huang}.} \bibinfo{year}{2025}\natexlab{}.
\newblock \showarticletitle{Memorag: Boosting long context processing with global memory-enhanced retrieval augmentation}. In \bibinfo{booktitle}{\emph{Proceedings of the ACM on Web Conference 2025}}. \bibinfo{pages}{2366--2377}.
\newblock


\bibitem[Shinn et~al\mbox{.}(2023)]%
        {shinn2023reflexion}
\bibfield{author}{\bibinfo{person}{Noah Shinn}, \bibinfo{person}{Federico Cassano}, \bibinfo{person}{Ashwin Gopinath}, \bibinfo{person}{Karthik Narasimhan}, {and} \bibinfo{person}{Shunyu Yao}.} \bibinfo{year}{2023}\natexlab{}.
\newblock \showarticletitle{Reflexion: Language agents with verbal reinforcement learning}.
\newblock \bibinfo{journal}{\emph{Advances in neural information processing systems}}  \bibinfo{volume}{36} (\bibinfo{year}{2023}), \bibinfo{pages}{8634--8652}.
\newblock


\bibitem[Shridhar et~al\mbox{.}(2021)]%
        {shridhar2021alfworld}
\bibfield{author}{\bibinfo{person}{Mohit Shridhar}, \bibinfo{person}{Xingdi Yuan}, \bibinfo{person}{Marc-Alexandre C{\^o}t{\'e}}, \bibinfo{person}{Yonatan Bisk}, \bibinfo{person}{Adam Trischler}, {and} \bibinfo{person}{Matthew Hausknecht}.} \bibinfo{year}{2021}\natexlab{}.
\newblock \showarticletitle{{ALFWorld: Aligning Text and Embodied Environments for Interactive Learning}}. In \bibinfo{booktitle}{\emph{Proceedings of the International Conference on Learning Representations (ICLR)}}.
\newblock
\urldef\tempurl%
\url{https://arxiv.org/abs/2010.03768}
\showURL{%
\tempurl}


\bibitem[Si et~al\mbox{.}(2026)]%
        {si2026eaglet}
\bibfield{author}{\bibinfo{person}{Shuzheng Si}, \bibinfo{person}{Haozhe Zhao}, \bibinfo{person}{Kangyang Luo}, \bibinfo{person}{Gang Chen}, \bibinfo{person}{Fanchao Qi}, \bibinfo{person}{Minjia Zhang}, \bibinfo{person}{Baobao Chang}, {and} \bibinfo{person}{Maosong Sun}.} \bibinfo{year}{2026}\natexlab{}.
\newblock \showarticletitle{A Goal Without a Plan is Just a Wish: Efficient and Effective Global Planner Training for Long-Horizon Agent Tasks}.
\newblock \bibinfo{journal}{\emph{ACL 2026}} (\bibinfo{year}{2026}), \bibinfo{pages}{13086--13113}.
\newblock


\bibitem[Song et~al\mbox{.}(2024)]%
        {song2024eto}
\bibfield{author}{\bibinfo{person}{Yifan Song}, \bibinfo{person}{Da Yin}, \bibinfo{person}{Xiang Yue}, \bibinfo{person}{Jie Huang}, \bibinfo{person}{Sujian Li}, {and} \bibinfo{person}{Bill~Yuchen Lin}.} \bibinfo{year}{2024}\natexlab{}.
\newblock \showarticletitle{Trial and error: Exploration-based trajectory optimization of LLM agents}. In \bibinfo{booktitle}{\emph{Proceedings of the 62nd Annual Meeting of the Association for Computational Linguistics (Volume 1: Long Papers)}}. \bibinfo{pages}{7584--7600}.
\newblock


\bibitem[Sun et~al\mbox{.}(2020)]%
        {sun2020lamol}
\bibfield{author}{\bibinfo{person}{Fan-Keng Sun}, \bibinfo{person}{Cheng-Hao Ho}, {and} \bibinfo{person}{Hung-Yi Lee}.} \bibinfo{year}{2020}\natexlab{}.
\newblock \showarticletitle{{LAMOL}: {LAnguage MOdeling} for Lifelong Language Learning}. In \bibinfo{booktitle}{\emph{International Conference on Learning Representations}}.
\newblock
\urldef\tempurl%
\url{https://openreview.net/forum?id=Skgxcn4YDS}
\showURL{%
\tempurl}


\bibitem[Team(2025)]%
        {qwen3technicalreport}
\bibfield{author}{\bibinfo{person}{Qwen Team}.} \bibinfo{year}{2025}\natexlab{}.
\newblock \bibinfo{title}{Qwen3 Technical Report}.
\newblock
\showeprint[arxiv]{2505.09388}~[cs.CL]
\urldef\tempurl%
\url{https://arxiv.org/abs/2505.09388}
\showURL{%
\tempurl}


\bibitem[Wang et~al\mbox{.}(2023a)]%
        {wang2024voyager}
\bibfield{author}{\bibinfo{person}{Guanzhi Wang}, \bibinfo{person}{Yuqi Xie}, \bibinfo{person}{Yunfan Jiang}, \bibinfo{person}{Ajay Mandlekar}, \bibinfo{person}{Chaowei Xiao}, \bibinfo{person}{Yuke Zhu}, \bibinfo{person}{Linxi Fan}, {and} \bibinfo{person}{Anima Anandkumar}.} \bibinfo{year}{2023}\natexlab{a}.
\newblock \showarticletitle{Voyager: An Open-Ended Embodied Agent with Large Language Models}. In \bibinfo{booktitle}{\emph{Intrinsically-Motivated and Open-Ended Learning Workshop @NeurIPS2023}}.
\newblock
\urldef\tempurl%
\url{https://openreview.net/forum?id=nfx5IutEed}
\showURL{%
\tempurl}


\bibitem[Wang et~al\mbox{.}(2023b)]%
        {wang2024dialogue}
\bibfield{author}{\bibinfo{person}{Huimin Wang}, \bibinfo{person}{Yunyan Zhang}, \bibinfo{person}{Yifan Yang}, \bibinfo{person}{Yefeng Zheng}, {and} \bibinfo{person}{Kam-Fai Wong}.} \bibinfo{year}{2023}\natexlab{b}.
\newblock \showarticletitle{Acquiring new knowledge without losing old ones for effective continual dialogue policy learning}.
\newblock \bibinfo{journal}{\emph{IEEE Transactions on Knowledge and Data Engineering}} \bibinfo{volume}{36}, \bibinfo{number}{12} (\bibinfo{year}{2023}), \bibinfo{pages}{7569--7584}.
\newblock


\bibitem[Wang et~al\mbox{.}(2024)]%
        {wang2023continualsurvey}
\bibfield{author}{\bibinfo{person}{Liyuan Wang}, \bibinfo{person}{Xingxing Zhang}, \bibinfo{person}{Hang Su}, {and} \bibinfo{person}{Jun Zhu}.} \bibinfo{year}{2024}\natexlab{}.
\newblock \showarticletitle{A comprehensive survey of continual learning: Theory, method and application}.
\newblock \bibinfo{journal}{\emph{IEEE transactions on pattern analysis and machine intelligence}} \bibinfo{volume}{46}, \bibinfo{number}{8} (\bibinfo{year}{2024}), \bibinfo{pages}{5362--5383}.
\newblock


\bibitem[Wang et~al\mbox{.}(2022)]%
        {wang2022scienceworld}
\bibfield{author}{\bibinfo{person}{Ruoyao Wang}, \bibinfo{person}{Peter Jansen}, \bibinfo{person}{Marc-Alexandre C{\^o}t{\'e}}, {and} \bibinfo{person}{Prithviraj Ammanabrolu}.} \bibinfo{year}{2022}\natexlab{}.
\newblock \showarticletitle{Scienceworld: Is your agent smarter than a 5th grader?}. In \bibinfo{booktitle}{\emph{Proceedings of the 2022 Conference on Empirical Methods in Natural Language Processing}}. \bibinfo{pages}{11279--11298}.
\newblock


\bibitem[Wang et~al\mbox{.}(2025)]%
        {wang2025awm}
\bibfield{author}{\bibinfo{person}{Zora~Zhiruo Wang}, \bibinfo{person}{Jiayuan Mao}, \bibinfo{person}{Daniel Fried}, {and} \bibinfo{person}{Graham Neubig}.} \bibinfo{year}{2025}\natexlab{}.
\newblock \showarticletitle{Agent Workflow Memory}. In \bibinfo{booktitle}{\emph{Forty-second International Conference on Machine Learning}}.
\newblock
\urldef\tempurl%
\url{https://openreview.net/forum?id=NTAhi2JEEE}
\showURL{%
\tempurl}


\bibitem[Xi et~al\mbox{.}(2024)]%
        {xi2024agentgym}
\bibfield{author}{\bibinfo{person}{Zhiheng Xi}, \bibinfo{person}{Yiwen Ding}, \bibinfo{person}{Wenxiang Chen}, \bibinfo{person}{Boyang Hong}, \bibinfo{person}{Honglin Guo}, \bibinfo{person}{Junzhe Wang}, \bibinfo{person}{Dingwen Yang}, \bibinfo{person}{Chenyang Liao}, \bibinfo{person}{Xin Guo}, \bibinfo{person}{Wei He}, \bibinfo{person}{Songyang Gao}, \bibinfo{person}{Lu Chen}, \bibinfo{person}{Rui Zheng}, \bibinfo{person}{Yicheng Zou}, \bibinfo{person}{Tao Gui}, \bibinfo{person}{Qi Zhang}, \bibinfo{person}{Xipeng Qiu}, \bibinfo{person}{Xuanjing Huang}, \bibinfo{person}{Zuxuan Wu}, {and} \bibinfo{person}{Yu-Gang Jiang}.} \bibinfo{year}{2024}\natexlab{}.
\newblock \bibinfo{title}{AgentGym: Evolving Large Language Model-based Agents across Diverse Environments}.
\newblock
\showeprint[arxiv]{2406.04151}~[cs.AI]


\bibitem[Yao et~al\mbox{.}(2022)]%
        {yao2022webshop}
\bibfield{author}{\bibinfo{person}{Shunyu Yao}, \bibinfo{person}{Howard Chen}, \bibinfo{person}{John Yang}, {and} \bibinfo{person}{Karthik Narasimhan}.} \bibinfo{year}{2022}\natexlab{}.
\newblock \showarticletitle{Webshop: Towards scalable real-world web interaction with grounded language agents}.
\newblock \bibinfo{journal}{\emph{Advances in Neural Information Processing Systems}}  \bibinfo{volume}{35} (\bibinfo{year}{2022}), \bibinfo{pages}{20744--20757}.
\newblock


\bibitem[Yao et~al\mbox{.}(2023)]%
        {yao2023react}
\bibfield{author}{\bibinfo{person}{Shunyu Yao}, \bibinfo{person}{Jeffrey Zhao}, \bibinfo{person}{Dian Yu}, \bibinfo{person}{Nan Du}, \bibinfo{person}{Izhak Shafran}, \bibinfo{person}{Karthik Narasimhan}, {and} \bibinfo{person}{Yuan Cao}.} \bibinfo{year}{2023}\natexlab{}.
\newblock \showarticletitle{{ReAct}: Synergizing Reasoning and Acting in Language Models}. In \bibinfo{booktitle}{\emph{International Conference on Learning Representations}}.
\newblock


\bibitem[Yen et~al\mbox{.}(2025)]%
        {yen2025slim}
\bibfield{author}{\bibinfo{person}{Howard Yen}, \bibinfo{person}{Yoonsang Lee}, \bibinfo{person}{Ashwin Paranjape}, \bibinfo{person}{Mengzhou Xia}, \bibinfo{person}{Thejas Venkatesh}, \bibinfo{person}{Jack Hessel}, \bibinfo{person}{Danqi Chen}, {and} \bibinfo{person}{Yuhao Zhang}.} \bibinfo{year}{2025}\natexlab{}.
\newblock \showarticletitle{Lost in the maze: Overcoming context limitations in long-horizon agentic search}.
\newblock \bibinfo{journal}{\emph{arXiv preprint arXiv:2510.18939}} (\bibinfo{year}{2025}).
\newblock


\bibitem[Zeng et~al\mbox{.}(2024)]%
        {zeng2024agenttuning}
\bibfield{author}{\bibinfo{person}{Aohan Zeng}, \bibinfo{person}{Mingdao Liu}, \bibinfo{person}{Rui Lu}, \bibinfo{person}{Bowen Wang}, \bibinfo{person}{Xiao Liu}, \bibinfo{person}{Yuxiao Dong}, {and} \bibinfo{person}{Jie Tang}.} \bibinfo{year}{2024}\natexlab{}.
\newblock \showarticletitle{Agenttuning: Enabling generalized agent abilities for llms}. In \bibinfo{booktitle}{\emph{Findings of the Association for Computational Linguistics: ACL 2024}}. \bibinfo{pages}{3053--3077}.
\newblock


\bibitem[Zhang et~al\mbox{.}(2025)]%
        {zhang2025memengine}
\bibfield{author}{\bibinfo{person}{Zeyu Zhang}, \bibinfo{person}{Quanyu Dai}, \bibinfo{person}{Xu Chen}, \bibinfo{person}{Rui Li}, \bibinfo{person}{Zhongyang Li}, {and} \bibinfo{person}{Zhenhua Dong}.} \bibinfo{year}{2025}\natexlab{}.
\newblock \showarticletitle{Memengine: A unified and modular library for developing advanced memory of llm-based agents}. In \bibinfo{booktitle}{\emph{Companion Proceedings of the ACM on Web Conference 2025}}. \bibinfo{pages}{821--824}.
\newblock


\bibitem[Zhang et~al\mbox{.}(2026)]%
        {zhang2026compose}
\bibfield{author}{\bibinfo{person}{Zheyuan Zhang}, \bibinfo{person}{Alvin Zhang}, \bibinfo{person}{Daniel Khashabi}, {and} \bibinfo{person}{Tianmin Shu}.} \bibinfo{year}{2026}\natexlab{}.
\newblock \showarticletitle{Continual Learning Mechanisms Compose for Long-Horizon Memorization}.
\newblock \bibinfo{journal}{\emph{arXiv preprint arXiv:2609.06986}} (\bibinfo{year}{2026}).
\newblock


\bibitem[Zhao et~al\mbox{.}(2024)]%
        {zhao2024expel}
\bibfield{author}{\bibinfo{person}{Andrew Zhao}, \bibinfo{person}{Daniel Huang}, \bibinfo{person}{Quentin Xu}, \bibinfo{person}{Matthieu Lin}, \bibinfo{person}{Yong-Jin Liu}, {and} \bibinfo{person}{Gao Huang}.} \bibinfo{year}{2024}\natexlab{}.
\newblock \showarticletitle{Expel: Llm agents are experiential learners}. In \bibinfo{booktitle}{\emph{Proceedings of the AAAI Conference on Artificial Intelligence}}, Vol.~\bibinfo{volume}{38}. \bibinfo{pages}{19632--19642}.
\newblock


\bibitem[Zheng et~al\mbox{.}(2024)]%
        {zheng2024synapse}
\bibfield{author}{\bibinfo{person}{Longtao Zheng}, \bibinfo{person}{Rundong Wang}, \bibinfo{person}{Xinrun Wang}, {and} \bibinfo{person}{Bo An}.} \bibinfo{year}{2024}\natexlab{}.
\newblock \showarticletitle{Synapse: Trajectory-as-exemplar prompting with memory for computer control}. In \bibinfo{booktitle}{\emph{International Conference on Learning Representations}}, Vol.~\bibinfo{volume}{2024}. \bibinfo{pages}{19036--19066}.
\newblock


\bibitem[Zhong et~al\mbox{.}(2024)]%
        {zhong2024memorybank}
\bibfield{author}{\bibinfo{person}{Wanjun Zhong}, \bibinfo{person}{Lianghong Guo}, \bibinfo{person}{Qiqi Gao}, \bibinfo{person}{He Ye}, {and} \bibinfo{person}{Yanlin Wang}.} \bibinfo{year}{2024}\natexlab{}.
\newblock \showarticletitle{Memorybank: Enhancing large language models with long-term memory}. In \bibinfo{booktitle}{\emph{Proceedings of the AAAI conference on artificial intelligence}}, Vol.~\bibinfo{volume}{38}. \bibinfo{pages}{19724--19731}.
\newblock


\bibitem[Zhou et~al\mbox{.}(2023)]%
        {zhou2024lats}
\bibfield{author}{\bibinfo{person}{Andy Zhou}, \bibinfo{person}{Kai Yan}, \bibinfo{person}{Michal Shlapentokh-Rothman}, \bibinfo{person}{Haohan Wang}, {and} \bibinfo{person}{Yu-Xiong Wang}.} \bibinfo{year}{2023}\natexlab{}.
\newblock \showarticletitle{Language agent tree search unifies reasoning acting and planning in language models}.
\newblock \bibinfo{journal}{\emph{arXiv preprint arXiv:2310.04406}} (\bibinfo{year}{2023}).
\newblock


\bibitem[Zhou et~al\mbox{.}(2024)]%
        {zhou2024webarena}
\bibfield{author}{\bibinfo{person}{Shuyan Zhou}, \bibinfo{person}{Frank~F Xu}, \bibinfo{person}{Hao Zhu}, \bibinfo{person}{Xuhui Zhou}, \bibinfo{person}{Robert Lo}, \bibinfo{person}{Abishek Sridhar}, \bibinfo{person}{Xianyi Cheng}, \bibinfo{person}{Tianyue Ou}, \bibinfo{person}{Yonatan Bisk}, \bibinfo{person}{Daniel Fried}, {et~al\mbox{.}}} \bibinfo{year}{2024}\natexlab{}.
\newblock \showarticletitle{Webarena: A realistic web environment for building autonomous agents}. In \bibinfo{booktitle}{\emph{International Conference on Learning Representations}}, Vol.~\bibinfo{volume}{2024}. \bibinfo{pages}{15585--15606}.
\newblock


\bibitem[Zhu et~al\mbox{.}(2025)]%
        {zhu2025knowagent}
\bibfield{author}{\bibinfo{person}{Yuqi Zhu}, \bibinfo{person}{Shuofei Qiao}, \bibinfo{person}{Yixin Ou}, \bibinfo{person}{Shumin Deng}, \bibinfo{person}{Shiwei Lyu}, \bibinfo{person}{Yue Shen}, \bibinfo{person}{Lei Liang}, \bibinfo{person}{Jinjie Gu}, \bibinfo{person}{Huajun Chen}, {and} \bibinfo{person}{Ningyu Zhang}.} \bibinfo{year}{2025}\natexlab{}.
\newblock \showarticletitle{Knowagent: Knowledge-augmented planning for llm-based agents}. In \bibinfo{booktitle}{\emph{Findings of the Association for Computational Linguistics: NAACL 2025}}. \bibinfo{pages}{3709--3732}.
\newblock


\end{thebibliography}


\appendix
\clearpage
\section{Continual Adaptation and Inference Procedures}
\label{app:algorithm}

Algorithms~\ref{alg:hamo-training} and~\ref{alg:hamo-inference} summarize the continual adaptation and task execution procedures used in our experiments. The notation follows Section~\ref{sec:method}. Coverage is computed within the incoming card collection. Selecting an anchor changes the training collection, while all experience cards are retained. Optional memory release is disabled and therefore omitted from the pseudocode. The replay sampler and knee-point detector are represented as subroutines.

\begin{algorithm}[H]
\caption{HaMo: Continual Adaptation}
\label{alg:hamo-training}
\small
\begin{algorithmic}[1]
\Require Stage datasets $D_1,\ldots,D_T$; frozen backbone $W_0$; initial adapter $\theta_0$
\Ensure Persistent memory $M_T$ and adapter $\theta_T$
\State $M_0 \gets \varnothing$; $H_0 \gets \varnothing$ \Comment{Historical trajectories}
\For{$t=1,\ldots,T$}
    \State $C_t \gets \{(g_i,o_i^0,\operatorname{Distill}(\tau_i)): \tau_i\in D_t\}$
    \State $M_t \gets M_{t-1}\cup C_t$
    \For{each card $i\in C_t$}
        \State $\mathcal{N}_3(i) \gets \operatorname{Top3}(i,C_t\setminus\{i\})$
    \EndFor
    \State $R_j \gets \{i:j\in\mathcal{N}_3(i)\}$ for each $j\in C_t$
    \State $S\gets [\,]$; $U\gets\varnothing$; $q\gets[0]$
    \While{an unselected candidate remains}
        \State $j^*\gets\arg\max_{j\notin S}|R_j\setminus U|$
        \If{$|R_{j^*}\setminus U|=0$}
            \State \textbf{break}
        \EndIf
        \State Append $j^*$ to $S$; $U\gets U\cup R_{j^*}$
        \State Append $|U|$ to $q$ \Comment{Coverage curve}
    \EndWhile
    \State $N_t^*\gets\operatorname{KneePoint}(q)$
    \State $A_t\gets\{\operatorname{SourceTrajectory}(S[m]):1\leq m\leq N_t^*\}$
    \State $B_t^{\mathrm{old}}\gets\operatorname{SampleReplay}(H_{t-1})$
    \State $B_t\gets A_t\cup B_t^{\mathrm{old}}$
    \For{each $\tau_i\in B_t$}
        \State $E_i\gets\operatorname{RetrieveOrMatch}(\tau_i,M_t)$
        \State Pair expert actions with $(g_i,o_i^0,h_{i,k},E_i)$
    \EndFor
    \State $\theta_t\gets\theta_{t-1}$
    \State Train $\theta_t$ on $\mathcal{L}_t$ for 3 epochs, learning rate $2\times10^{-4}$
    \State Keep $W_0$ frozen; retain the same LoRA adapter
    \State $H_t\gets H_{t-1}\cup D_t$
\EndFor
\end{algorithmic}
\end{algorithm}

The greedy selection in Algorithm~\ref{alg:hamo-training} is written explicitly for clarity; the implementation uses lazy priority-queue updates to obtain the same marginal-coverage selection. The historical collection $H_t$ denotes the pool available for replay, rather than an additional trainable model. At the first stage, the replay collection is empty. The training loss is the experience-conditioned behavioral cloning objective in Section~\ref{sec:model-adaptation}.

\begin{algorithm}[H]
\caption{HaMo: Experience-Conditioned Inference}
\label{alg:hamo-inference}
\small
\begin{algorithmic}[1]
\Require Task $g$, initial observation $o^0$, memory $M_t$, adapter $\theta_t$, frozen backbone $W_0$
\State $E\gets\operatorname{Retrieve}((g,o^0),M_t)$
\State Construct initial dialogue from $(g,o^0,E)$; set $h_0\gets[o^0]$
\For{$k=0,\ldots,49$}
    \State Sample $a_k\sim\pi_{\theta_t}(\cdot\mid g,o^0,h_k,E)$
    \Statex \hspace{\algorithmicindent}\hspace{\algorithmicindent}using decoding temperature $0.7$
    \State $(o^{k+1},r_k,d_k)\gets\operatorname{EnvironmentStep}(a_k)$
    \State $h_{k+1}\gets h_k\mathbin{\Vert}(a_k,o^{k+1})$
    \If{$d_k$ indicates termination}
        \State \textbf{break}
    \EndIf
\EndFor
\State \Return Recorded trajectory and environment score
\end{algorithmic}
\end{algorithm}

Inference queries use only the task and initial observation. Retrieved experience remains guidance: subsequent actions also depend on the evolving interaction history. No expert solution or future observation is required to construct the inference query.

\section{Implementation and Experimental Settings}
\label{app:settings}

Experiments use NVIDIA A800 GPUs with 80\,GB device memory. At each adaptation stage, we train the selected anchor trajectories together with historical replay for three epochs at a learning rate of $2\times10^{-4}$. Only the LoRA adapter is updated, with rank $r=16$ and scaling parameter $\alpha=32$; the backbone remains frozen. The same adapter is carried forward across stages rather than reinitialized or replaced by a task-specific adapter. Table~\ref{tab:implementation-settings} summarizes the settings.

\begin{table}[H]
\centering
\caption{HaMo implementation and evaluation settings.}
\label{tab:implementation-settings}
\small
\begin{tabular}{@{}p{0.47\columnwidth}p{0.47\columnwidth}@{}}
\toprule
\textbf{Setting} & \textbf{Value} \\
\midrule
GPU hardware & NVIDIA A800, 80\,GB \\
Training epochs per stage & 3 \\
Learning rate & $2\times10^{-4}$ \\
LoRA rank $r$ & 16 \\
LoRA scaling parameter $\alpha$ & 32 \\
Backbone parameters & Frozen \\
Adapter management & One adapter, updated sequentially \\
Inference temperature & 0.7 \\
Maximum interaction turns & 50 per episode \\
Coverage-selection neighbors & 3, excluding the query card \\
Anchor budget & Stage-specific coverage-curve knee \\
Memory release & Disabled; all cards retained \\
\bottomrule
\end{tabular}
\end{table}

\paragraph{Stochastic agent inference.}
We set the inference temperature to $0.7$ to retain diversity in action generation after trajectory-based training. Interactive tasks may require the agent to adjust its decisions when the current state differs from a retrieved example; sampling is intended to reduce overly deterministic execution patterns and allow alternative actions. This is a decoding choice, not a training regularizer or a guarantee against overfitting. We do not attribute the reported gains specifically to temperature, since a temperature ablation is not included.

\paragraph{Evaluation protocol.}
Each episode ends when the environment terminates or after 50 interaction turns. We use the native task scores and report them on a 0--100 scale, retaining partial credit for ScienceWorld and WebShop. The within-domain protocol uses four sample-disjoint ScienceWorld training stages and evaluates the full test set after each stage. Cross-domain evaluation follows the task orders described in Section~\ref{sec:experimental-setting} and measures performance on domains encountered so far.






\end{document}